# PERFORMANCE ANALYSIS OF MULTICLASS SUPPORT VECTOR MACHINE CLASSIFICATION FOR DIAGNOSIS OF CORONARY HEART DISEASES

Wiharto[1,2], Hari Kusnanto[2] and Herianto[3]

[1]Department of Informatic, Sebelas Maret University,Indonesia,
[2]Department of Biomedical Engineering (BME),Gadjah Mada University, Indonesia and
[3]Department of Mechanical and Industrial Engineering,Gadjah Mada University, Indonesia

## ABSTRACT

*Automatic diagnosis of coronary heart disease helps the doctor to support in decision making a diagnosis. Coronary heart disease have some types or levels. Referring to the UCI Repository dataset, it divided into 4 types or levels that are labeled numbers 1-4 (low, medium, high and serious). The diagnosis models can be analyzed with multiclass classification approach. One of multiclass classification approach used, one of which is a support vector machine (SVM). The SVM use due to strong performance of SVM in binary classification. This research study multiclass performance classification support vector machine to diagnose the type or level of coronary heart disease. Coronary heart disease patient data taken from the UCI Repository. Stages in this study is preprocessing, which consist of, to normalizing the data, divide the data into data training and testing. The next stage of multiclass classification and performance analysis. This study uses multiclass SVM algorithm, namely: Binary Tree Support Vector Machine (BTSVM), One-Against-One (OAO), One-Against-All (OAA), Decision Direct Acyclic Graph (DDAG) and Exhaustive Output Error Correction Code ( ECOC). Performance parameter used is recall, precision, F-measure and Overall accuracy. The experiment results showed that the multiclass SVM classification algorithm with the algorithm BT-SVM, OAA-SVM and the ECOC-SVM,gave the highest Recall in the diagnosis of type or healthy level with a value above 90%, precision 82.143% and 86.793% F-measure,. For all kinds of algorithms, except binary OAA-SVM algorithm gave the highest recall 0.0% for the type or level of sick-high and sick-serious, and ECOC- SVM algorithm gave the highest recall 0.0% for sick-medium and sick-serious. While the type or level other, the performance of recall, precision and F-measure between 20% - 30%,. The conclusion that can be drawn is that the approach to the multiclass classification algorithm BT-SVM, OAO-SVM, DDAG-SVM to diagnosis the type or level of coronary heart disease provides better performance, than the binary classification approach.*

## 1.INTRODUCTION

Coronary heart disease is a narrowing of the arteries disease. The disease causes high mortality in a group of non-communicable diseases. Prevention and early detection of the disease is very important to suppress the rate of deaths from the disease. Referring to the dataset UCI Repository of coronary heart disease, heart disease grouped into 4 types or levels of a given symbol 1-4 [1]. According to [2] and [3], coronary heart disease classified into 4 types or levels, namely low, medium, high and serious [2] [3]. Based on these groupings, then the diagnosis to determine the type or level of coronary heart disease is very important. Good diagnosis of the type or level of coronary heart disease, will be used to provide appropriate action. Diagnosis of type or level of coronary heart disease can use multiclass classification approach.

Research conducted by [4] that studied of the use of computational intelligence algorithms for the diagnosis of coronary heart disease. The study did multiclass classification conversion to binary classification. Selection of the classification, due to the classification method SMO (Support Vector Machine Optimization) has a good ability in binary classification [4]. Classification is done by creating a positive and negative labels for each level. If we would classify a low level, then the low-level data is given a positive label, while another level in addition to the low level given the negative label. The method makes use of the classification into binary classification, because it only detects positive or negative. Results of testing with 10-fold crossvalidation method shows good accuracy, but very low for a true positive (TP) and F-measure for low-level, high, medium and serious. Computational intelligence algorithms tested in the study were naive Bayesian, SMO, IBK, AdaaBoostM1, J48 and PART.

Subsequent research conducted by [5]. The study uses a classification approach and computational intelligence algorithms similar to research done by [4]. Research conducted by [5] using 10-fold cross-validation with iterations as much as 100 times, while testing the significance of differences between computational intelligence algorithms using t-test. The test results showed that the SMO gives better performance of the other algorithm. Other similar studies have also been conducted by [6]. This study is similar to that done before, by using the binary classification approach. This study adds randomize the process prior to the 10-fold cross validation. The processes are carried out as many as 100 times, while the end result is the average yield of 100 times the [6]. Multiclass clasification conversion approach to binary classification also performed by [7], its research emphasize banchmarking feature selection methods, computational intelligence algorithm C4.5 and Naive Bayesian.

The next approach used is multiclass classification approachment. Research has been done by [8], which is an intelligence system for predictive type or level of coronary heart disease using methods Weighted Associative Classifier (WAC). The method is able to provide improved performance, when compared with the other associative classification methods [8]. Following research conducted by [9], is a research proposes a hybrid classification with a genetic algorithm, a modified k-NN and back propagation neural network. Testing of the system by using one dataset coronary heart disease cleveland UCI Repository. The test results demonstrate the accuracy reached 62.1%. The result is better than multiclass classification algorithms like naive Bayesian tested and k-NN [9]. Furthermore, research conducted by [10] and [11], which used neural network to classify heart disease [10] [11]. In the study of heart disease is classified into four classes, namely normal person, a first stroke, a second stroke and end of life. In the study conducted by [11], in addition to using 13 variables from the dataset cleveland UCI Repository, also using variable plus namely obesity and smoking. The addition of these variables make better neural network performance in terms of accuracy.

The studies that have been conducted by [4], [5] and [6] showed that the SMO algorithm gives good performance. The downside of these studies did not use the approach of multiclass classification Support Vector Machine. Method of Support Vector Machine (SVM) is a powerful binary classification. SVM also developed for the multiclass clacification. The classification developed by the two approaches. They are the optimization of multi-class classification problems and classification compiled from a binary SVM [12]. Research [12] describes some of the settlement of multi-class classification, using a combination of a number of binary SVM. These methods are One-Against-All (OAA), One-Against-One (OAO) and Directed Acyclic Graph (DAG). The [13] also proposed multiclass SVM such as Binary Decision Tree archtecture (Binary Tree SVM). Beside the [13], the [14] [15] [16] also developed multiclass SVM Error Correction Output Code (ECOC).

Referring to the previous studies, this research will be to analyze the performance of multi-class SVM classification for the diagnosis of type or level of coronary heart disease. Classification algorithm used is Binary Tree SVM, OAO, OAA, DDAG and ECOC. Performance parameters in the analysis is the recall, precision and F-measure. Research using cleveland dataset UCI Repositor

## 2.MATAERIAL AND METHOD

### 2.1.Material

This study used a dataset of coronary heart disease from UCI Repository [1]. Coronary heart disease data distributed into five types or levels. They are healthy, low-pain, pain-medium, ill-high and ill-serious. Coronary heart disease data from the UCI Repository obtained from the collection of Robert Detrano, MD, Ph.D., at VA Medical Center. The amount of data consists of 303 intance the data, the number of parameters 14, with one parameter as an indication of the level of heart disease with a scale of 0-4. Influencing parameters is shown in detail in Table 1.

Table 1.Parameters that influence coronary heart disease.

| No | Parameters | Description |
|---|---|---|
| 1 | Age | Age (continues data) 29 to 77 |
| 2 | Sex | Sex : 1: man, 0: woman |
| 3 | Cp | Chest pain type : <br> 1 : Typical Angina <br> 2 : atypical angina <br> 3 : non-anginal pain <br> 4 : asymtomatic |
| 4 | Trestbps | Diastolic blood pressure  (mm Hg) |
| 5 | Chol | Cholesterol in mg/dl |
| 6 | Fbs | Fasting blood suger >120 mg/dl,  1 : true, 0 : false |
| 7 | Restecg | Resting ECG :  0 : Normal, 1 : ST-T abnormal, 2 : Left V. Hypertrophy |
| 8 | Thalach | Maximum heart rate achieved |
| 9 | Exang | Exercise induced angina (1=yes, 0=no) |
| 10 | Oldpeak | ST depression induced byexercise relative to rest |
| 11 | Slope | The slope of the peak exercise ST Segment : 1 : ups loping, 2: flat, 3 : downsloping |
| 12 | Ca | Number of major vessels colored by flouroscopy (0-3) |
| 13 | Thal | Defect type : 3 : Normal, 6 : fixed defect, 7 : reversable defect |
| 14 | Num | Heart disease (0-4) : 0=Healthy,1= low,2= medium, 3=high, 4=serious. |

## 2.2.Method

This research method is divided into several stages, namely preprocessing, data sharing, training, testing and performance analysis. Preprocessing stages, is to normalize the data, which bringing the data into a certain scale. Normalization method used is using the Min-Max.

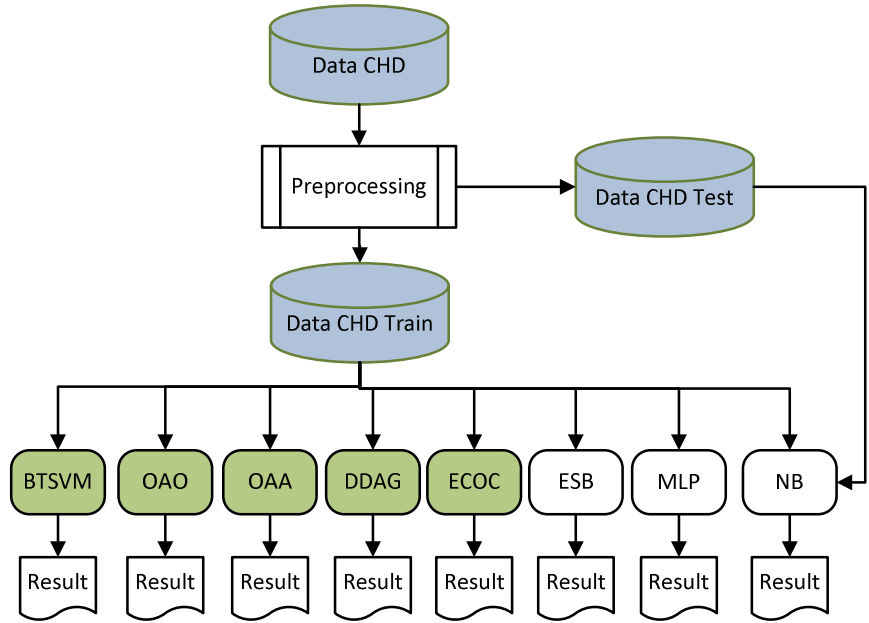

Figure 1. Coronary heart disease diagnosis system with multiclass SVM

The next stage after the preprocessing is to divide the data into two groups of data. They are training and testing. The distribution of the data can be shown in Table 2.

Table 2. Dataset distribution Coronary Heart Disease

| Level | Training | Testing |
|---|---|---|
| **Healthy** | 114 | 50 |
| **Sick-Low** | 38 | 17 |
| **Sic- Medium** | 20 | 16 |
| **Sick-High** | 24 | 11 |
| **Sick-Serious** | 10 | 3 |
| | 206 | 97 |

The next process is to training. Training done using some multiclass SVM algorithm are: BT, OAO, OAA, DDAG, ECOC and as a comparison algorithm multiclass Multi Layer Perceptron (MLP), Naive Bayesian (NB) and ESB (ensemble AdaaBoostM1) [17] [18]. After the training, then performed testing using data that has been prepared. The parameters measured in testing (testing) is recall (sensitifity), precision, F-measure and overall accuracy. Explanation for each parameter with reference to Table 3 is as follows

    a.  Precision or confidence, is the number of positive samples were classified correctly categorized as positive divided by the total data sample testing is classified as positive.

$$Precision(confidence) = \frac{TP}{TP+FP} \tag{1}$$

b. Recall or sensitifity, is the number of positive samples were classified correctly category divided by the total positive samples in the data were categorized testing positive

$$Recall(sensitifity) = \frac{TP}{TP+FN} \tag{2}$$

c. F-Measure (F1), is the harmonic mean of Recall and Precision

$$F1 = 2 \cdot \frac{Precision*Recall}{Precision+Recall} \tag{3}$$

d. Overall Accuracy, expressed in the following equation

$$Overall\ Accuracy = \frac{TP + TN}{TP + TN + FP + FN} \tag{4}$$

Table 3. Confusion Matrics

| Actual Class | | Prediction Class | |
|---|---|---|---|
| | | Positif | Negatif |
| | Positif | TP | FN |
| | Negatif | FP | TN |

## 3. RESULT

Diagnosis of type or level of coronary heart disease is a problem that can be solved with some approaches of multiclass classification. Research results multiclass SVM classification performance analysis can be shown each level. Figure 2 is a comparison of the performance of SVM multiclass in diagnosing healthy, or in any other sense diagnosis healthy or sick. BTSVM algorithm performance, OAA, OAO, DDAG, ECOC provide almost the same performace, even BTSVM, OAA and the ECOC provide recall performance above 90%. The precision and F-measure reached above 80%. A good performance, when compared to the two levels of diagnostic accuracy using SVM algorithm with a binary classification approach, as it typically involves in previous studies [19] [20] [21] [4].

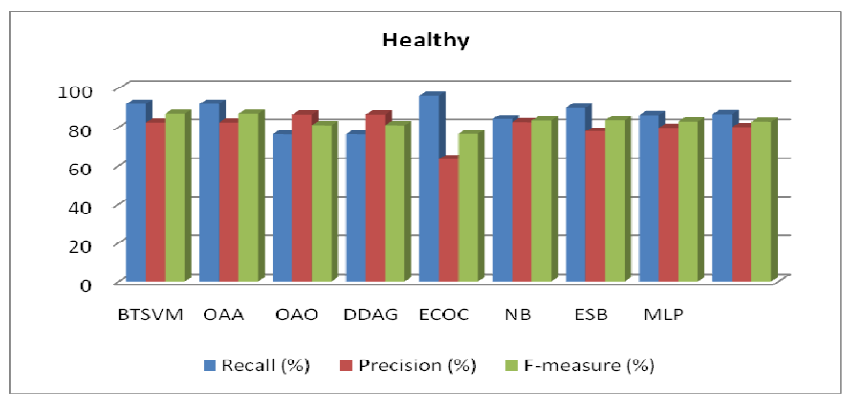

Figure 2: Performance diagnosis system at the level of Healthy

The results of further research, shown in Figure 3-6. The picture shows the performance of the type or level of pain-low, ill-medium, ill-high and ill-serious with multiclass SVM algorithm, Ensemble, naive Bayesian and multi-layer perceptron. Performance on the type or level sick-low is very low even to below 40% better recall, precision and F-measure. Only one algorithm, that is naive Bayesian recall that could reach above 50%. similar with that the type or level of sick-medium, sick-high and sick-serious, the value of the three variables of performance below 50%, which gives performance ECOC only recall, precision and F-measure above 60% for the type or level of sick-High.

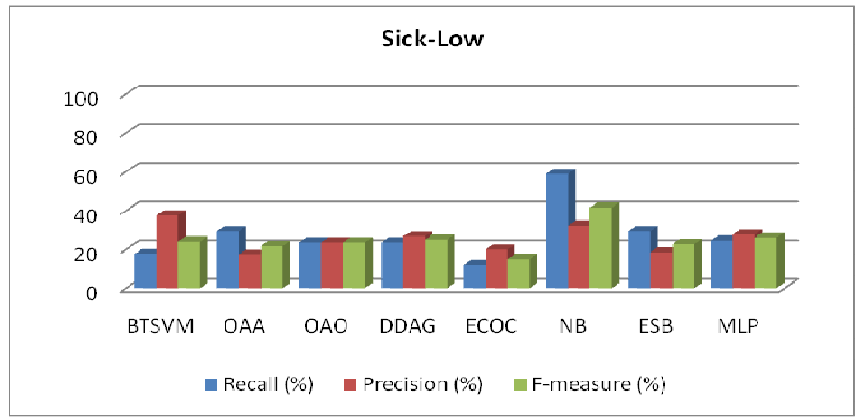

Figure 3: Performance diagnosis system at the level of Sick-Low

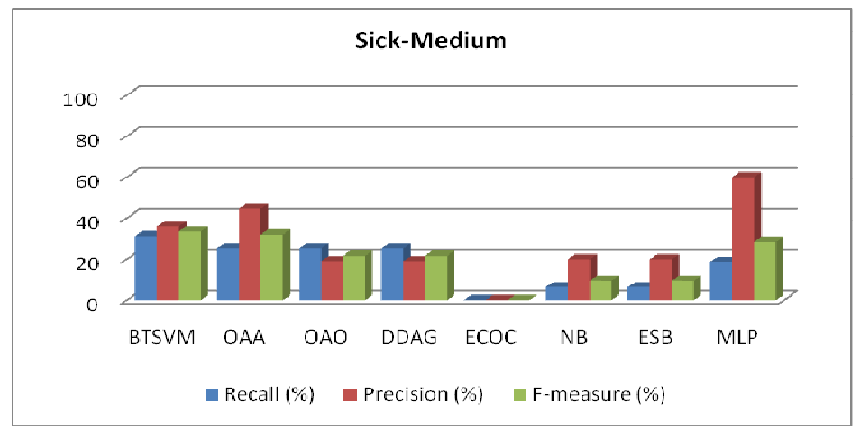

Figure 4: Performance diagnosis system at the level of Sick-Medium

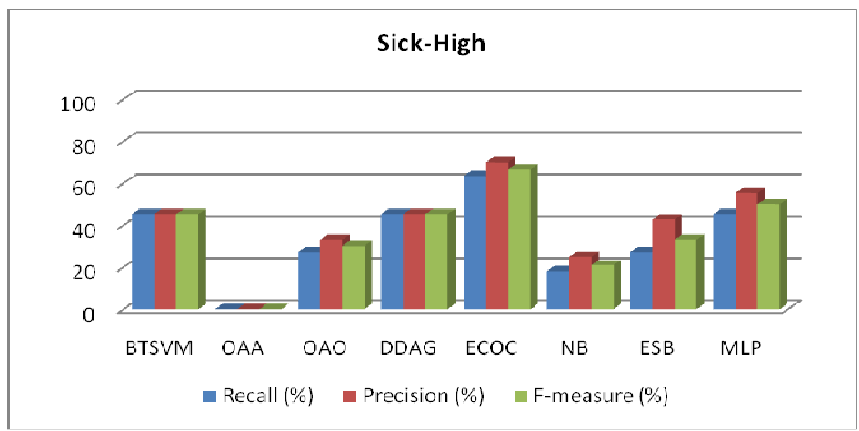

Figure 5: Performance diagnosis system at the level of Sick-High

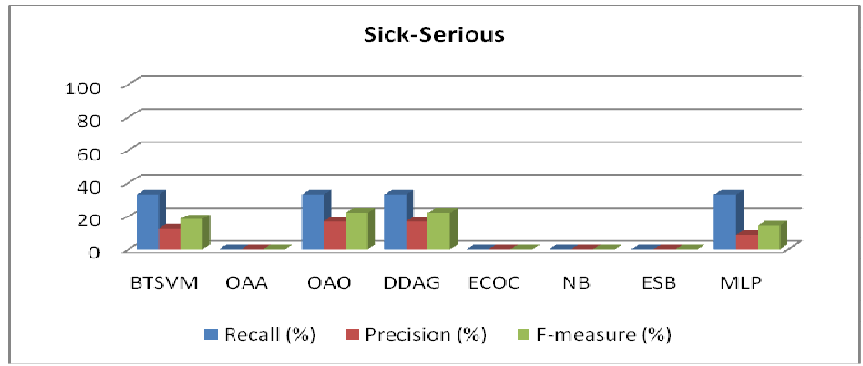

Figure 6: Performance diagnosis system at the level of Sick-Serious

The results of the next study is overall calculation accuracy of each algorithm. Fully shown in Figure 7. Overall accuracy is generated no more than 70% for all algorithms, even just BTSVM algorithm which has overall accuracy above 60%.

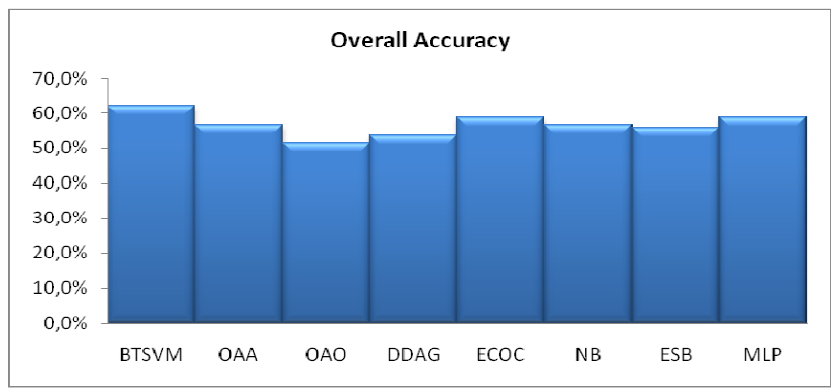

Figure 7.Overall Accuracy Multiclass Classification

## 4.DISCUSSIONS

Discussions in this study is that we are going to compare some research on the diagnosis of coronary heart disease ever undertaken, using either a multiclass classification approach or multiclass classification conversion to binary classification. First, the research that has been done by [4]. The study uses a binary classifications approach, with reference to the strong performance of the classification of SMO. Results of these studies was in good accuracy performance, for each type or level, but has a value of TP (True Positive) and F-measure is very low. The condition shows that high due to the high accuracy True Negatives (TN), or the system has a good ability in classifying negative data, but less well in classifying the positive data. In addition, due to the labeled amount of data is more negative than positive labeled. The results of a similar study was also conducted by [6], which in this study has a high accuracy, but TP and F-measure low. When compared with the research using SVM classification mutliclass, we take the same performance parameters, namely the F-measure, show the performance of SVM multiclass give better results from research conducted by [4] and [6] for type or level sick-low, sick-medium, sick-high and sick-serious. Similar research has been done by [5], F-measure results better. The repair is done by adding the number of iterations in the test using 10-fold cross validation. Multiclass SVM classification performance comparison with previous studies, the full shown in Table 4.

Table 4. Comparison of F-measure binary classification approach with multiclass

| Author | Classification Approach | Methode | F-Measure (%) | | | | |
|---|---|---|---|---|---|---|---|
| | | | Healthy | Sick-Low | Sick-Medium | Sick-High | Sick-Serious |
| Nahar et,al [4] | Binary | SMO | **81,800** | 0,000 | 0,000 | 0,000 | 0,000 |
| Prabowo et.al[6] | Binary | SMO | **84,200** | 0,000 | 0,100 | 3,500 | 0,000 |
| Akrami et.al [5] | Binary | SMO* | **87,000** | **90,000** | **93,000** | **94,000** | **98,000** |
| This Study | Multiclass | BT | **86,793** | 24,000 | 33,333 | 45,455 | 18,182 |
| | Multiclass | OAA | **86,793** | 21,739 | **32,000** | 0,000 | 0,000 |
| | Multiclass | OAO | 80,851 | 23,529 | 21,622 | 30,000 | **22,222** |
| | Multiclass | DDAG | 80,851 | **25,000** | 21,622 | **45,455** | 22,223 |
| | Multiclass | ECOC | 76,200 | 14,800 | 0,000 | **66,700** | 0,000 |

\* with iterations as much as 100 times

The next comparison, are multiclass classification method. If we look at the performance of each of multiclass SVM, the BT-SVM algorithm, OAO-SVM and SVM-DDAG recall that delivers performance, precision, F-measure relatively stabil in any type or level. It is different for OAA-ECOC-SVM and SVM there is some level of value 0. If we compare with SMO (SVM Optimization) with a binary classification approach made [4] and [6], SVM multiclass have more stable performance in any type or level of coronary heart disease. The downside of multiclass classification is only able to give a good performance in diagnosing healthy, and not good for the other type or level.

The next comparison is with previous studies that use multiclass classification approach. First, research done by [8]. The study proposed a system of intelligence to predict heart disease using Weighted Associative Classifier (WAC) [8]. The system was tested using multiple datasets, one of the data is heart disease from UCI Repository [1]. The test results showed overall system accuracy is still relatively low, that below 60% or below multiclass SVM performance. The study did not measure performance for each type or level, so it can not be compared to each type or level with multiclass SVM. Using the same concept, the [9] proposed a hybrid system. The system is tested using several datasets, including UCI dataset Repository. The test results provide performance overall accuracy of 62.1%. Hybrid system proposed in the study using a combination of some of our algorithms, namely Genetic algorithm, neural netwok and modified k-NN. Performance generated in these studies was slightly better than the performance of SVM multiclass TB-SVM. Multiclass SVM research for the diagnosis of coronary heart defect, resulting in overall performance 61.86% accuracy (BT-SVM) and the lowest 51.546% (OAO-SVM). The poor performance due to the low capacity of the system in diagnosing the type or level of low, medium, high and serious of coronary heart disease. Comparison multilcass SVM classification system performance with some previous studies can be shown in Table 5.

Table 5. Overall Accuracy comparison multiclass classification

| Author | Methode | Overall Accuracy (%) |
|---|---|---|
| Soni et.al [8] | WAC | 57,75 |
| | CBA | **58,28** |
| | CMAR | 53,64 |
| | CPAR | 52,32 |
| Salari et.al [9] | Hybrid (ANN+GA+Modification k-NN) | **62,10** |
| This Study | **BT-SVM** | **61,86** |
| | OAA-SVM | 56,70 |
| | OAO-SVM | 51,546 |
| | DDAG-SVM | 53,608 |
| | ECOC-SVM | **58,763** |
| | NB | 56,701 |
| | ESB | 55,67 |
| | MLP | **58,763** |

The next discussion is knowing the significance differences between the performance of particular recall for the type or level of coronary heart disease. Significance testing the performance difference between the type or level recall of coronary heart disease using t-test, with a 95% confidence level (p-value <0.005). Significance test results are shown in Table 6. From these results indicate that the performance difference between healthy recall the type or level of low, medium, high and serious produce p-value <0.005 (with a 95% confidence level), it means there are significant differences in the performance of recall. If we connect with the amount of data used for training, the amount of data is more healthy than the type or level of low, medium, high and serious. Referring to the research that has been done by [22], resulting in a problem of data imbalance. A machine learning will yield a good prediction accuracy classification of the class of data training with large numbers of members, while the number of class members have a little less good accuracy [22].

Tabel 6. Significance test results using a t-test

| No | Comparison | p-value |
|---|---|---|
| 1 | **Healthy-Low** | **0,0000000240** |
| 2 | **Healthy-Medium** | **0,0000000005** |
| 3 | **Healthy-High** | **0,0000032178** |
| 4 | **Healthy-Serious** | **0,0000000354** |
| 5 | Low-Medium | 0,0677226456 |
| 6 | Low-High | 0,2222116996 |
| 7 | Low-Serious | 0,1024362015 |
| 8 | **Medium-High** | **0,0280130119** |
| 9 | Medium-Serious | 0,4727597556 |
| 10 | **High-Serious** | **0,0430968053** |

The significance of test results for type or level low, medium, high and serious p-value> 0.05, meaning that there is not significant a differences. If we look at the data training used to have the same relative amount of difference between the low-medium, low-high and low-serious, so its performance is relatively the same. Type or level medium compared to high and serious, for medium-high there was significant difference p-value <0.05. This is because the data type or level high more than the type or level of medium. Referring to the research conducted by [22], a better performance. Furthermore, for type or level high-serious also showed a significant difference, and if we see the amount of data training occurred is also relatively large differences.

# 5.CONCLUSIONS

Research multiclass SVM classification performance analysis for the diagnosis of coronary heart disease, some conclusions can be drawn. First, BT-SVM multiclass classification, OAO-SVM and SVM-DDAG provide better performance than the classification of the binary classification approach. Secondly, BT-SVM multiclass classification, OAO-SVM and DDAG-SVM provide a relatively stabile performance for all levels, but there is a difference between the level of performance that is significant at the level with minimal training data. Third, the occurrence of imbalanced data on coronary heart disease data for sick-low levels, sick-medium, sick-high and sick-serious, resulting in low performance of the system.